\documentclass[lettersize,journal]{IEEEtran}
\pdfoutput=1
\usepackage{amsmath,amsfonts}
\usepackage{algorithmic}
\usepackage{algorithm}
\usepackage{array}
\usepackage[caption=false,font=normalsize,labelfont=sf,textfont=sf]{subfig}
\usepackage{textcomp}
\usepackage{stfloats}
\usepackage{url}
\usepackage{verbatim}
\usepackage{graphicx}
\usepackage{cite}
\usepackage{xcolor}
\usepackage[implicit=false]{hyperref}
\usepackage{color}
\usepackage{caption}

\begin{document}

\title{Deep Active Learning with Structured Neural Depth Search}

\author{Xiaoyun~Zhang,
        Yiping~Xie,
        Jianwei~Zhang     

\thanks{Xiaoyun Zhang and Yiping Xie contributed equally to this work.
\emph{(Corresponding author: Jianwei Zhang.)}}
\thanks{Xiaoyun Zhang, Yiping Xie, and Jianwei Zhang are with the College of Computer Science, Sichuan University,
        Chengdu 610065, China (e-mail: finkannets@gmail.com; wingsxyp@gmail.com; zhangjianweio@stu.edu.cn).}
}

\markboth{IEEE TRANSACTIONS ON NEURAL NETWORKS AND LEARNING SYSTEMS}%
{Shell \MakeLowercase{\textit{et al.}}: Deep Active Learning with a Structured Neural Depth Search}


\maketitle

\begin{abstract}
Previous work optimizes traditional active learning~(AL) processes with incremental neural network architecture search~(Active-iNAS) based on data complexity change, which improves the accuracy and learning efficiency. However, Active-iNAS trains several models and selects the model with the best generalization performance for querying the subsequent samples after each active learning cycle. The independent training processes lead to an insufferable computational budget, which is significantly inefficient and limits search flexibility and final performance. To address this issue, we propose a novel active strategy with the method called structured variational inference~(SVI) or structured neural depth search~(SNDS) whereby we could use the gradient descent method in neural network depth search during AL processes. At the same time, we theoretically demonstrate that the current VI-based methods based on the mean-field assumption could lead to poor performance. We apply our strategy using three querying techniques and three datasets and show that our strategy outperforms current methods.
\end{abstract}


\section{Introduction}
\IEEEPARstart{D}{eep} learning~(DL) has recently received promising results on various tasks due to the powerful learning capacities of deep neural networks~(DNNs). DNNs always have complex structures and their performances highly rely on a large number of annotated samples. However, collecting a large labeled dataset is rather challenging in practice, especially for the medical domains where the annotations are expensive and slow to produce~\cite{lu2017deep}. Active learning~(AL) focuses on alleviating the reliance on expensive tremendous annotated datasets~\cite {ren2021survey}. It assumes that different samples have different contributions to the DNN performance, and tries to select the most deterministic samples to construct the training set.

\begin{figure*}[h]
\begin{center}
\centerline{\includegraphics[width=2\columnwidth]{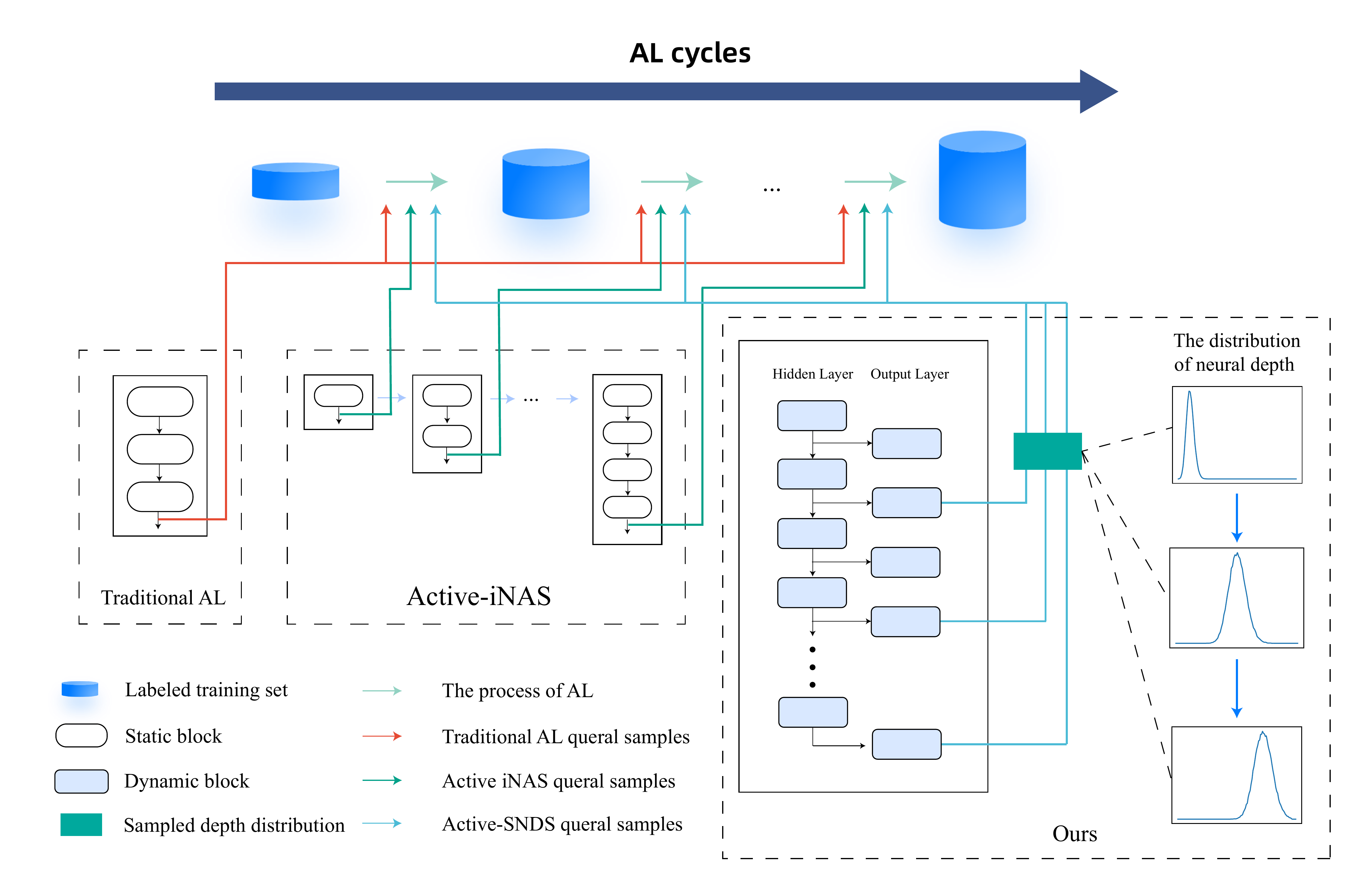}}
\caption{Illustrations of previous methods and our proposed method in an active learning process. Left: \textbf{Traditional AL}, where a pre-designed fixed model is used throughout the whole active learning process. Mid: \textbf{Active-iNAS}, where networks with different depths are trained in the active learning process and the current optimal network architecture is selected for active learning. Right:  \textbf{Active-SNDS}~(Ours), where a single model continuously searches for new neural network depth distributions during active learning and updates the network depth by stacking new basic blocks of distributions, using the hybrid weighted output of different depths of different networks for active learning.}
\label{fig_framework}
\end{center}
\end{figure*}

Traditional AL methods mostly leverage a fixed model to query samples, which is ineffective to fit the increasing scale of the labeled training set. At the early stage of an AL method, an overparameterized model is unlikely to achieve a good generalization performance with a small-scale training set. In contrast to previous AL methods with fixed architectures, Geifman et al.~\cite{geifman2019deep} proposed an active learning method with incremental neural architecture search~(Active-iNAS) to dynamically increase the neural capacity during the AL cycles. As shown in Fig.~\ref{fig_framework}, Active-iNAS starts with a small neural capacity at the early stages. Then, an incremental neural architecture search~(NAS)~\cite{zoph2016neural, pham2018efficient, liu2018darts, cai2018proxylessnas} is performed to monotonically non-decrease the neural capacity. It maintains a well-generalized model to ensure query effectiveness. Active-iNAS can overwhelmingly outperform traditional AL methods with fixed architectures. However, Active-iNAS trains several models and selects the model with the best generalization performance for querying the subsequent samples after each AL cycle. The independent training processes of Active-iNAS lead to an insufferable computational budget, which is significantly inefficient and limits search flexibility and final performance.

This paper summarizes the above approaches into a new problem setting, called Neural Depth Search~(NDS) in AL context, automatically searching an optimal neural architecture depth for the given task during active learning processes. NDS is a crucial complement to NAS for its ability to adjust neural capacity automatically. In this problem setting, we aim to answer the following key question: \textit{How to design an NDS method in AL processes to improve search efficiency and promote model performance?}


To address the aforementioned issue, Variational Inference~(VI) is utilized to learn the neural architecture depth uncertainty aiming at achieving better search flexibility and final performance, which approximates a parameterized posterior over the architectural depth~\cite{blei2017variational}. The variational parameters~\cite{blei2017variational, zhang2018advances} are optimized by minimizing the Kullback-Liebler divergence~(KL-divergence) between the approximated posterior and the true one. At last, the approximated posterior can reflect the probabilistic reasoning over neural architecture depth for different tasks. Furthermore, we propose a novel AL strategy with the method called Structured Variational Inference~(SVI) or Structured Neural Depth Search~(SNDS) based on VI, which makes the neural weights dependent on the architecture depth and improves the fidelity of the posterior approximation and final search performance. The differences of these methods are shown in Fig. ~\ref{fig_framework}. Specifically, we implement SVI using Pseudo-uniform sampling at the depth for training the sharing weights. 



We call our learning approach Active-SNDS and apply Active-SNDS using three querying techniques~(random, uncertainty entropy~\cite{krishnakumar2007active} and coreset~\cite{sener2017active,geifman2017deep}) and three datasets~(CIFAR10~\cite{krizhevsky2009learning}, CIFAR100~\cite{krizhevsky2009learning}, and MNIST~\cite{lecun-mnisthandwrittendigit-2010}). The findings are exciting: 1) SVI can search more reasonable neural network depth according to the current dataset scale and complexity than iNAS and current VI-based method; 2) Active-SNDS outperforms Active-iNAS and VI-based stragety using different query methods in active learning processes.

Our contributions can be summarized as follows:
\begin{itemize}
\item We theoretically demonstrate that the mean-field assumption of current VI-based methods can cause the rich-get-richer problem, i.e., the shallow networks would dominate the search and be outputted as the results.
\item We proposed the SVI method that can restore the mean-field assumption and search for more reasonable neural network depth according to the current dataset scale and complexity than Active-iNAS and VI-based method;
\item We propose the Active-SNDS method address problems in Active-iNAS. Extensive experiments demonstrate Active-SNDS outperforms Active-iNAS and VI-based stragety in active learning processes.
\end{itemize}

The remainder of this paper is organized as follows. Section II reviews the related works of AL and VI-based NDS and compares them with our method. We present preliminaries for AL and VI-based NDS problems in Section III and propose our SVI and Active-SNDS methods in Section IV. Section V demonstrate the experimental results of NDS and AL. Finally, Section VI concludes our paper.

\section{Related Work}
\subsection{Deep Active Learning}
A standard AL method is initialized with an unlabeled sample pool and a manually designed model~\cite{ren2021survey}. The model follows a certain strategy to query the most valuable samples from the pool in cycles. The bundle of samples are labeled by an oracle and merged into the training set. The labeled training set is then used for training the model again. This process repeats until exhausting the label budget reaches the expected model performance. 

The query strategy, which refers to how to select the samples to be labeled, is key to the performance of an AL method~\cite{ren2021survey}. Among them, the most popular strategy is the uncertainty-based method, which ranks all the samples with a metric called uncertainty~\cite{wang2016cost}. A great uncertainty indicates a higher selecting priority. The Density-based methods attempt to select the core set that represents the distribution of the entire dataset~\cite{sener2017active, geifman2017deep}. However, previous AL methods mostly leverage a fixed model to query samples, which is ineffective to fit the increasing scale of the labeled training set. At the early stage of an AL method, an overparameterized model is unlikely to achieve a good generalization performance with a small-scale training set.

To mitigate this problematic aspect, the discussion of architecture optimization in active learning was considered within the context of deep neural models. Huang et al.~\cite{huang2015efficient} demonstrated that active learning performance can be improved using a proper choice of (fixed) hyperparameters in the context of linear models. Geifman et al.~\cite{geifman2019deep} proposed Active-iNAS to dynamically increase the neural capacity during the AL cycles, which further enhances active learning flexibility and accuracy. Our work differs from the above work by combining VI to learn the neural architecture depth uncertainty instead of increasing the basic block thereby it could achieve better search flexibility and final performance.

\subsection{The current VI-based methods for NDS}
There are only a few studies related to NDS. 
Dikov et al.~\cite{dikov2019bayesian} proposed to estimate the neural architecture width and depth through BNN. Antorán et al.~\cite{antoran2020variational} proposed to search the depth of residual networks in the efficient one-shot NAS framework, where the neural weights and architecture are jointly learned. Antorán et al.~\cite{antoran2020depth} estimated the depth uncertainty through the probabilistic reasoning over a sequential structure of feed-forward networks. 
Nazaret et al.~\cite{nazaret2022variational} propose a novel VI method to approximate the posterior over the neural weights and depth of an infinitely deep network.

Most of the current methods~\cite{dikov2019bayesian, antoran2020variational, antoran2020variational} approximate the architecture depth posterior with VI based on the mean-field assumption~\cite{blei2017variational, zhang2018advances}, where the neural weights and depth variables are independent. The mean-field assumption can limit the approximation fidelity and introduce the rich-get-richer problem, i.e., the shallow networks would dominate the search. Different from the previous method, a structured variational inference study is proposed to restore the mean-field assumption, significantly improving the search effectiveness. Same to the proposed method, Nazaret et al.~\cite{nazaret2022variational} also utilize the SVI to impose the dependence between the neural weights and depth. The difference is that we theoretically demonstrate how the rich-get-richer problem is caused by the mean-field assumption. And Pseudo-uniform sampling method is proposed to implement the SVI.

\section{Preliminaries}
In this section, we present the preliminaries including deep active learning, searching the optimal neural depth during active learning, and VI.

\subsection{Deep Active Learning}
In active learning setting, a large unlabeled data pool $U = \{\mathcal{X}, \mathcal{Y}\}^{N_u}$ can be obtained, where $\mathcal{X}$ denotes the sample space, $\mathcal{Y}$ denotes the label space but it is unknown before labeled by oracle. Given an initialized labeled dataset $L = \{x_i, y_i\}^{N_{init}}_{i=1}$, we can train a deep model $f\in\mathcal{F}: \mathcal{X} \rightarrow \mathcal{Y}$, where $\mathcal{F}$ is hypothesis space of the model. Then, we cyclically sample new bunch of samples $\mathcal{B^*}$ from $U$ to be labeled:

\begin{equation}\label{query_process}
\mathcal{B}^*=\underset{\mathcal{B} \subseteq \mathcal{U}}{\arg \max } \quad a(\mathcal{B}, f)
\end{equation}

where $a$ is the query strategy. $\mathcal{B^*}$ would be labeled by oracle and added to the $L$. The query cycle repeats until reaching an ideal model performance or the label budget is exhausted. The goal of AL is to obtain a higher model performance with a lower label budget $|L|$.

From the AL query process Eq.~(\ref{query_process}), we could find the model's generation performance and its induced query effectiveness are critical to the final performance of AL. However, according to the statistical learning theory~\cite{geifman2019deep, vapnik1999overview, he2019data}, with probability at least $1 - \delta$, the generation gap can be bounded as:
\begin{equation}\label{VC_theory}
R(f) - \hat{R}_{L}(f) \leq O( \sqrt{ \frac{ d_{VC}\log(N_t/d_{VC}) - \log{\delta} }{N_t} } ), 
\end{equation}
where $R(f)$ is the expected risk on the unseen sample such as $U$, $\hat{R}_{L}(f)$ is the empirical risk on the training set $L$, $d_{VC}$ is the VC-Dimension of $\mathcal{F}$, and $N_t$ is the dynamical sample number of $L$ that increases linearly with the AL cycles. Let the term under the root, $\frac{ d_{VC}\log(N_t/d_{VC}) - \log{\delta} }{N_t}$, denoted by $G$, we have its partial derivative on $d_{VC}$:

\begin{equation}
\frac{\partial G}{\partial d_{V C}}=\frac{1}{N_t}\left(\log N_t / d_{V C}-1\right)
\end{equation}

When $\log{N_t/d_{VC}}=1$, we can always get the minimum of the generation gap bound. This means that $d_{VC}$ should grow linearly with the increasing $N_t$. As neural capacity is positively related to $d_{VC}$~\cite{bartlett2019nearly}, Geifman et al.~\cite{geifman2019deep} proposed to monotonically non-decrease the neural capacity during the AL process.  

	

\subsection{Searching the optimal neural depth during active learning}
A truncated Poisson distribution has been used to model the neural depth~\cite{nazaret2022variational}, denoted as $\bar{\mathcal{P}}(\lambda, d_{min}, d_{max})$, where $\lambda$ is the mean and also the variational parameter, and $d_{min}$ and $d_{max}$ denote the minimum and maximum of the available neural depths. We can obtain the probability at each depth $d$ :
\begin{equation}\label{discrete_approximation}
\bar{\mathcal{P}}(X=d)=\frac{{\mathcal{P}}(X=d)}{\sum_{j=d_{min}}^{d_{max}}{\mathcal{P}}(X=j) }.
\end{equation}
	
We set  $d_{min}$ to be 1 and $d_{max}$ is calculated by the following formula:
\begin{equation}\label{d_max}
d_{max} = m(q^{0.95}({\lambda})), 
\end{equation}
where $\left\{q^{\delta}({\lambda}) \mid q \in \mathcal{P}\right\}$ represents the distribution Poisson($\lambda$) truncated to its $\delta$-quantile and $m(q) := max\{\ell|q(\ell) > 0\}$.

\subsection{Variational Inference}
Neural depth search is challenging for two factors. On the one hand, deep neural networks are always over-parameterized and would like to overfit the training data~\cite{zhang2021understanding}. The model's performance on the training set can not determine the optimal architecture depth. On the other hand, navigating the depth search through the generalization performance is impractical because it requires a large validation set for stable evaluation, which is not feasible for the annotation-expensive domains like medical images~\cite{zhang2020automated}. The current mean-field VI-based studies~\cite{dikov2019bayesian, antoran2020depth, antoran2020variational} inherit the advantages of BNN such as modeling uncertainty and eliminating overfitting. 

Given an observed data set $\mathcal{D} = \{x_i, y_i\}^N_{i=1}$ and a $L$-layer network, we consider two kinds of variables, i.e., the neural weights $w$ and the architecture depth $d$. We define a prior distribution over the architecture depth $p(d, w)$ and a likelihood $p(\mathcal{D}|d, w)$ are defined. The posterior can be computed through exact inference $p(d, w|\mathcal{D}) = \frac{p(d, w)p(\mathcal{D}|d, w)}{p(\mathcal{D})}$. However, the posterior is usually difficult to calculate because the evidence $p(\mathcal{D})$ is intractable. 

Variational inference approximates a parameterized surrogate distribution $q(d, w)$ to the posterior $p(d, w|\mathcal{D})$. Previous VI-based methods are all implicitly under the mean-field assumption, where the neural weights $w$ and the architecture depth $d$ are independent. $q(d, w)$ can be factorized as:
\begin{equation}\label{meanfield}
q(d, w) = q_{\lambda}(d)q_{\mu, \sigma}(w),
\end{equation}
where $\lambda$ is the parameter of the architecture depth distribution, and the neural weights are defined as Gaussian distribution: $w \sim \mathcal{N}(\mu, diag(\sigma^2))$, $\mu, \sigma \in \mathbb{R}^{|w|}$. In the following context, the variational parameters~\cite{blei2017variational, zhang2018advances} $\lambda$, $\mu$, and $\sigma$ are sometimes omitted for easy presentation. Then, VI aims at finding an optimal member from the variational family $q^*(\cdot)$ closest to the exact posterior $p(d, w|\mathcal{D})$ in Kullback-Leibler divergence~(KL-divergence)~\cite{blei2017variational}. 

\begin{equation}\label{kldivergence}
\begin{split} 
&KL(q(d, w)\|p(d, w|\mathcal{D}))  \\
&= -ELBO   + \log{p(\mathcal{D})} ,
\end{split}
\end{equation}
where ELBO is an acronym for the evidence lower bound, which is a tractable surrogate objective of KL-divergence. As the term $\log{p(\mathcal{D})}$ does not depend on the variational parameters, maximizing the ELBO is equivalent to minimizing the KL-divergence. VI turns the inference problem into an optimization problem. In the NDS problem setting, we focus on the optimization of the architecture depth parameter $\lambda$. Following~\cite{dikov2019bayesian}, a point estimation is treated over the neural weights $w$. ELBO is approximated by the Monte Carlo~(MC) sampling over $q_{\lambda}(d)$ and maximum a posterior~(MAP) estimation over $w$. Following~\cite{antoran2020depth}, we directly optimize neural weights $w$ instead of their variational parameters $\mu$ and $\sigma$. As a result, $\lambda$ and $w$ can be optimized simultaneously using stochastic gradient descent:
\begin{equation}\label{KL_opti}
\begin{split} 
&\ell(\lambda,  w) =  -ELBO  \\
&= \mathbb{E}_{q_{\lambda}(d)}[
-\sum_{i=1}^{N}\log{p(y_i;f(x_i, w_{1:d}))  } ]
\\ 
&~~~~+KL(q_{\lambda}(d)\|p(d)) + KL(q(w)\|p(w)),
\end{split}
\end{equation}

where $B$ is the batch size and $f(x_i, w_{1:d})$ denotes the output of the $d$-depth network that only have the neural weights from $1$-th to $d$-th layer. For the last line of Eq.~(\ref{KL_opti}), the first term is the expectation of cross-entropy~(CE) loss under the surrogate distribution over the architecture depth, the second term can be calculated through MC sampling over $d$, and the third term has a closed form solution and can be computed analytically~\cite{dikov2019bayesian}. 


\section{Proposed Methods}
In this section, we theoretically demonstrate that the mean-field assumption of previous VI-based methods can cause the rich-get-richer problem and propose our SVI and Active-SNDS methods to improve the search flexibility and final performance.

\subsection{The Dark Side of Mean-Field Assumption}
The mean-field assumption makes it easy to capture any marginal density of the variables and estimate the ELBO~\cite{blei2017variational}. However, it can limit the fidelity of the posterior approximation and cause the rich-get-richer problem, i.e., the shallow layers' weights have more chance to be trained and the shallow architectures tend to be outputted. 
\begin{equation}\label{darkside}
\begin{split} 
&\frac{\partial \ell(\lambda,  w)}{\partial w}  =\frac{\partial }{\partial w}( \sum_{d}^{}{q_{\lambda}(d)} \ell_{ce}(D, d, w_{1:d}) + \ell_{kl}),
\end{split}
\end{equation}
where $\ell_{ce}(D, d)=\sum_{i=1}^{N}-\log{p(y_i;f(x_i, w_{1:d}))  } $ is the CE loss of the $d$-depth network on $D$, and it only relies on the neural weights from $1$-th to $d$-th layer $w_{1:d}$. $\ell_{kl}$ denotes the sum of the two KL terms and it is independent with $d$. $\ell_{kl}$ is usually much smaller than the CE loss that is calculated with the sum mode. 

We have the gradients of $l$-th layer:
\begin{equation}\label{eachlayer}
\begin{split} 
\frac{\partial \ell(\lambda,  w)}{\partial w_l}   = \frac{\partial }{\partial w_l}( \sum_{d=l}^{L}{q_{\lambda}(d)} \ell_{ce}(D, d, w_{1:d}) + \ell_{kl}).
\end{split}
\end{equation}
If we consider that $\ell_{ce}(D, d)$ under different depths are equal and $\ell_{kl}$ is much smaller compared to the CE loss, the ratio of the last layer's gradients to the first layer's could be roughly estimated as $\frac{\partial \ell(\lambda,  w)}{\partial w_L} / \frac{\partial \ell(\lambda,  w)}{\partial w_1}   = q_{\lambda}(L)$. In other words, the shallow layers will converge much faster than the latter ones. Then, the CE loss of the shallow networks would be smaller than the deep ones.

\begin{figure}[ht]
	\vskip 0.2in
	\begin{center}
		\centerline{\includegraphics[width=\columnwidth]{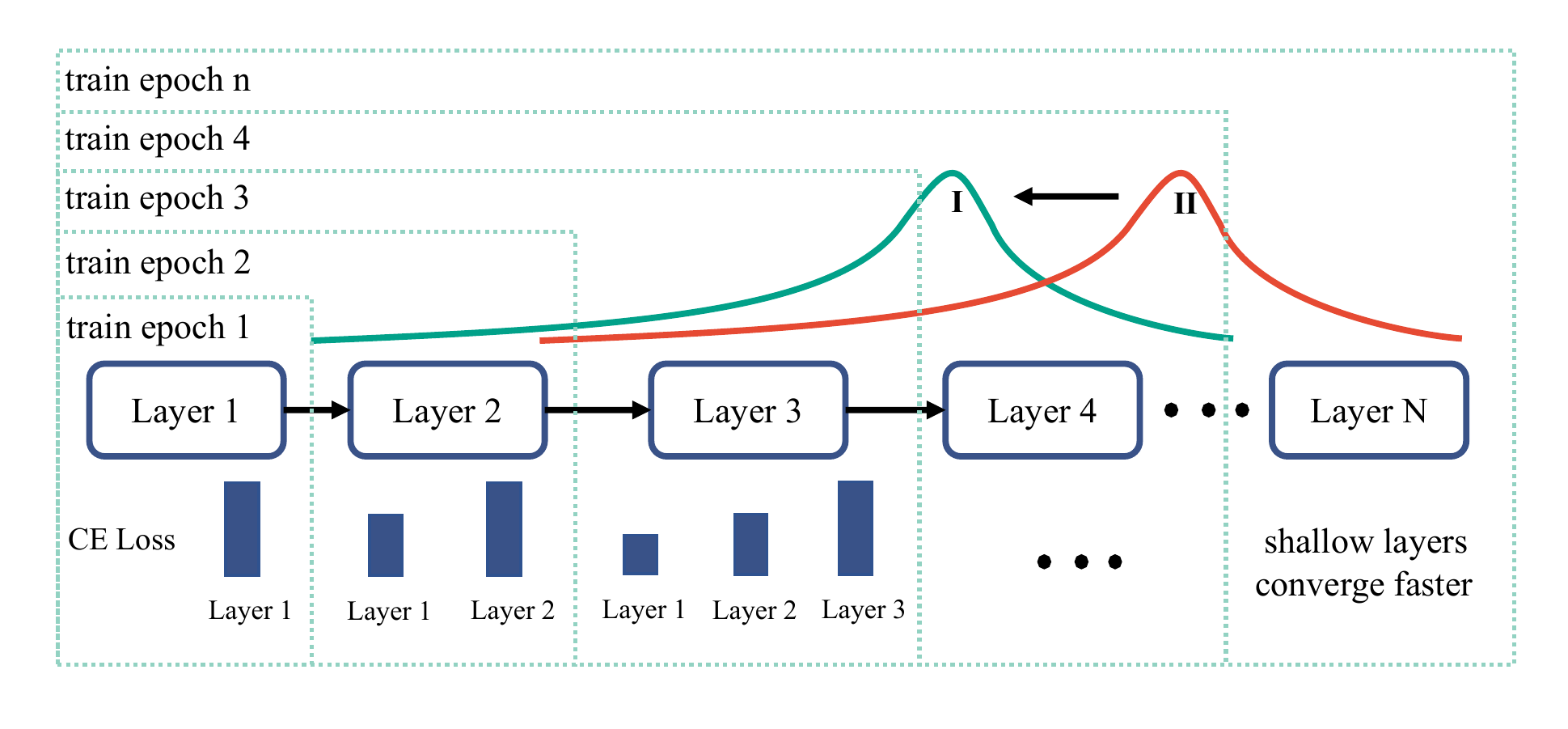}}
		\caption{As training progresses, the shallow network has been training so that the CE loss decreases more (as shown in the bar), so converges faster, and the network tends to give the shallow layer a larger sampling probability (curve I in green) to make the overall convergence of the network faster than to give the deep layer (curve II in orange).}
		\label{fig:darkside}
	\end{center}
	\vskip -0.2in
\end{figure}

Moreover, we consider the learning of $\lambda$:
\begin{equation}\label{lambda}
\begin{split} 
&\frac{\partial \ell(\lambda,  w)}{\partial \lambda}  =\frac{\partial }{\partial \lambda}( \sum_{d}^{}{q_{\lambda}(d)} \ell_{ce}(D, d, w_{1:d}) + \ell_{kl}).
\end{split}
\end{equation}
As the CE loss of the shallow networks is smaller, the sampling probability $q_{\lambda}(d)$ over the shallow network would converge to be larger. This entails the rich-get-richer problem~\cite{adam2019understanding} and navigates the search toward shallow networks as shown with an intuitive example in Fig.~\ref{fig:darkside}.

\subsection{Neural Depth Search with Structured Variational Inference}
\begin{figure}[ht]
	\vskip 0.2in
	\begin{center}
		\centerline{\includegraphics[width=\columnwidth]{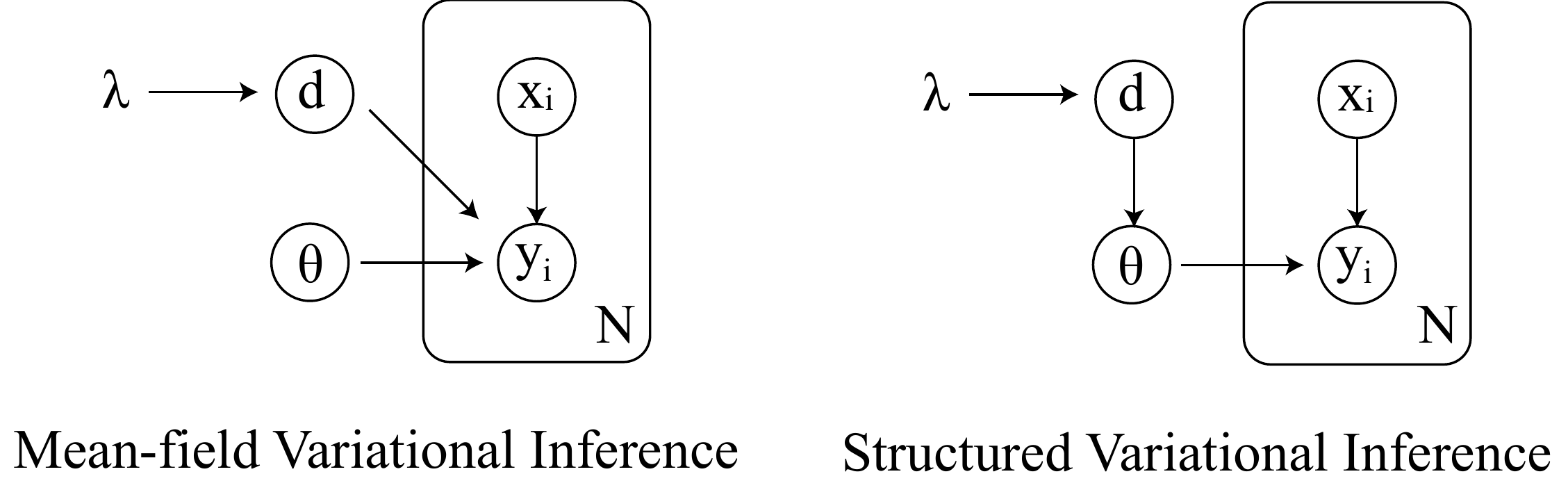}}
		\caption{The graphical models of the traditional mean-field VI and the proposed SVI. $x_i$ and $y_i$ denote the sample and label from a dataset. $\theta$ denotes the neural weights. $d$ and $\lambda$ denote the architecture depth and its variational parameters. In the mean-field VI, the neural weights and architecture depth are independent. In SVI, the neural weights are conditioned on the architecture depth.}
		\label{fig_svi}
	\end{center}
	\vskip -0.2in
\end{figure}

The mean-field assumption in previous VI-based methods limits the fidelity of the posterior approximation and introduces local optima in neural depth search. In this paper, we propose to relax the independence between the neural weights and the architecture depth through structured variational inference.Fig.~\ref{fig_svi} gives an intuitive example of the mean-field VI and SVI. As a result, the weights of networks with different depths can be customized, which can greatly improve the performance of variational inference in NDS.

Specifically, SVI models the neural weights to be dependent on the architecture depth in Eq.~(\ref{meanfield}):
\begin{equation}\label{strocutecVI}
q(d, w) = q_{\mu, \sigma}(w|d)q_{\lambda}(d),
\end{equation}
The loss function over $\lambda$ and $w$ in Eq.~({\ref{KL_opti}}) can be reformulated as:
\begin{equation}\label{svi}
\begin{split} 
&\ell(\lambda,  w)=\mathbb{E}_{q(w|d)q(d)}[\log{q(w|d)}+\log{q(d)} \\
&~~~~-\log{p(w|d)}-\log{p(d)} - \log{p(D|d, w)}] \\
&=KL(q_{\lambda}(d)\|p(d)) + \mathbb{E}_{q_{\lambda}(d)}[KL(q(w|d)\|p(w|d))\\  &-\sum_{i=1}^{N}\log{p(y_i;f(x_i, w|d))}].   
\end{split}
\end{equation}

It could be found that the main differences between Eq.~(\ref{svi}) and Eq.~(\ref{KL_opti}) are: 1) the KL term of the neural weights are under the expectation over $q_{\lambda}(d)$; 2) the neural weights are conditioned on the depth. 

Specifically, we propose a Pseudo-uniform sampling method to implement Eq.~(\ref{svi}). Motivated by some one-shot NAS methods which have already an explored effective way to eliminate the unfair advantages of early dominant operations by uniformly sampling candidate architectures for training the sharing weights~\cite{zhang2021one, guo2020single}, we propose to use early uniform sampling of the networks with each depth to address the drawback of mean-field assumption in NDS. The training of the sharing weights could be optimized as:
\begin{equation}\label{uniform}
w^* =\operatorname*{arg\,min} \mathbb{E}_{d\sim\Gamma }[\ell_{ce}(D, d)],
\end{equation}
where $\Gamma$ denotes the uniform distribution over the depth choice set. In this paper, our model is a growing unbounded depth neural network~\cite{nazaret2022variational}, so we cannot simply use uniform sampling as in the one-shot NAS method. In order to make the whole sampling uniform in the process of model growth, we propose a pseudo-uniform sampling method that uses the inverse of the frequency of sampling in each layer as the probability of sampling in the next training. And then substitute $w^*_{1:d}$ as $w|d$ in Eq.~(\ref{svi}).

\subsection{Active Learning with Structured Neural Depth Search}
The Active-SNDS technique is described in Algorithm \ref{activesnds} and works as follows. Given the unlabeled data $U$ and the initial labeled data $L$, it is used to train the model and obtain the unlabeled data for annotation in the active learning process. The whole process consists of two nested loops. The outer loop is the active learning cycle, and the inner loop is the network training cycle. The depth parameter $\lambda$ and network parameters $w_{1:d_{max}}$ are trained using pseudo-random sampling in the first 1/3 of the network training cycles and mean-field VI loss functions in the last 2/3 of the cycles. During the depth update, basic blocks are generated and stacked by the layer generator $l$ to join the current network model~\cite{nazaret2022variational}. When the network training period is over, the acquisition function is used to collect $b$ labeled data from $U$ for annotation and join the labeled data $L$. When the active learning cycle is complete, exit the outer loop. More detailed algorithm implementation will be given in Section V.

\begin{algorithm}
   \caption{Active Learning with Structured Neural Depth Search~(Active-SNDS).}
   \label{activesnds}
\begin{algorithmic}
   \STATE {\bfseries Input:} unlabeled data $U$; labeled data $L$; budget b; layer generators $l$;
   \STATE Initialize: $\lambda$
   \STATE $layers = []$
   \FOR{$i=1$ {\bfseries to} $T_{i}$}
   \FOR{$j=1$ {\bfseries to} $T_{j}$}
   \STATE Compute $d_{max}$
   \WHILE{$k:=|layers| < d_{max}$} 
   \STATE Add new layer $l(k+1)$ to $layers$ 
   \STATE Initialize $w_{k+1}$
   \ENDWHILE
   \IF{$j <= T_{j}/3$}
      \STATE Compute $\ell(\lambda,  w)$ with Pseudo-uniform sampling
   \ELSE
      \STATE Compute $\ell(\lambda,  w)$ without Pseudo-uniform sampling
   \ENDIF
   \STATE Compute gradients $\nabla_{\lambda, w_{1:d_{max}}} \ell(\lambda,  w)$
   \STATE Update $\lambda,  w_{1:d_{max}}$
   \ENDFOR
   \STATE $B = \phi$
   \WHILE{$|B| < b$}
   \STATE Add acquire($U$) to $B$ and sub $B$ from $U$
   \ENDWHILE
   \STATE Add $B$ to $L$
   \ENDFOR
\end{algorithmic}
\end{algorithm}

\section{Experiments}
In this section, we will show the performance comparison of our proposed methods Active-SNDS, mean-field VI and Active-iNAS, as well as two fixed networks Resnet-18 and Resnet-34 in depth search and active learning, and make a comprehensive analysis of the corresponding results. 

\subsection{Datasets}
We used three datasets to evaluate our proposed method, namely MNIST, CIFAR-10 and CIFAR-100. 

\begin{itemize}
\item \textbf{MNIST:} The MNIST database was constructed from NIST's Special Database 3 and Special Database 1 which contain binary images of handwritten digits. It has a training set of 60,000 examples and a test set of 10,000 examples including 0-9 ten handwritten digit classes. And the digit images we used in the MNIST set were originally selected and experimented with by Chris Burges and Corinna Cortes using bounding-box normalization and centering, so the size of each image is 28*28 .
\item \textbf{CIFAR-10:} The CIFAR-10 is labeled subsets of the 80 million tiny images dataset. They were collected by Alex Krizhevsky, Vinod Nair, and Geoffrey Hinton. This dataset consists of 60000 32*32 colour images in 10 classes, with 6000 images per class. There are 50000 training images and 10000 test images.
\item \textbf{CIFAR-100:} The CIFAR-100 is also labeled subset of the 80 million tiny images dataset. It is just like the CIFAR-10, except it has 100 classes containing 600 images each. There are 500 training images and 100 testing images per class.
\end{itemize}

In depth search related experiments, we used all training samples in the corresponding dataset; In the active learning related experiments, the total number of training samples we ended up using was only a part of three datasets, with the final number of samples used on MNIST being 2300 and the final sample size of 45000 on CIFAR-10 and CIFAR-100. 

\subsection{Implementation Details}
\begin{figure}[ht]
	\vskip 0.2in
	\begin{center}
		\centerline{\includegraphics[width=\columnwidth]{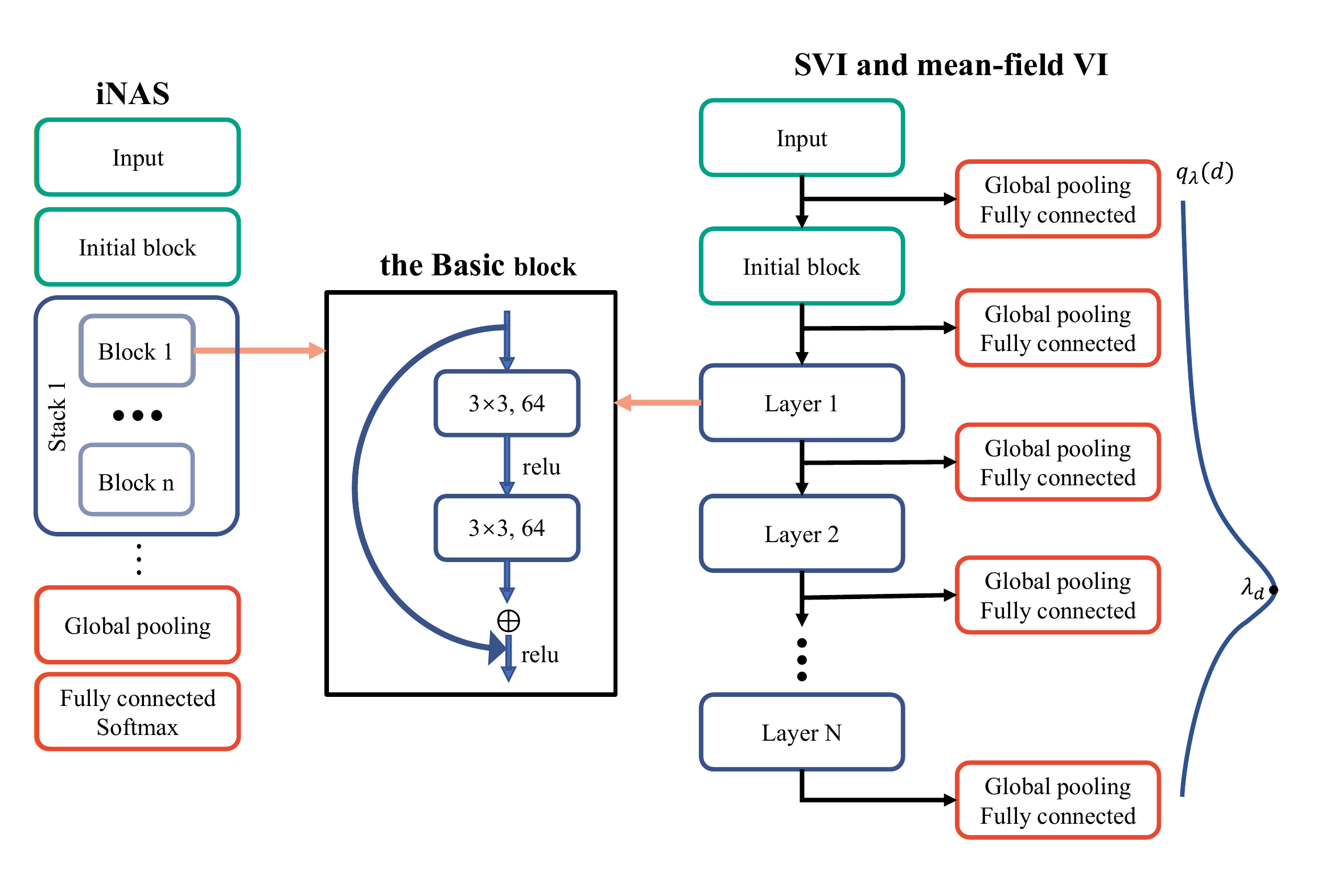}}
		\caption{Network architectures under methods iNAS, SVI and mean-field VI. For the network model searched by iNAS, the output is directly the output of the deepest network. For the network model searched by mean-field VI and SVI, the output is the weighted sum of the predicted output of each layer and their respective probabilities corresponding to the truncated Poisson distribution $q_\lambda(d)$.}
		\label{models_pdf}
	\end{center}
	\vskip -0.2in
\end{figure}

\textbf{Model Details.} In the paper, we used two search strategies, mean-filed VI and iNAS, and two fixed network structures, Resnet-18 and Resnet-34, to compare with our proposed method SVI. Among them, the search space of mean-filed VI and SVI is an infinite depth network, while the search space of iNAS is set to $A(B, 1, 1)$ to $A(B, 12, 5)$ ($A(B, i, j)$: the network structure represented in iNAS consists of stacks; $B$ is a fixed neural block; $i$ is the number of blocks in each stack; $j$ is the number of stacks). In the experiments, based on the Resnet architecture~\cite{he2016deep}, we used the basic block as the basic building block of the search strategies and treated it as a layer in the network depth. Specifically, the basic block contains two convolutional layers of size 3*3 followed by a batch normalization and then a ReLU activation. A residual connection is added before the activation of the second convolutional layer. The overall network architectures of these search methods are shown in Fig. \ref{models_pdf}.

\begin{itemize}
\item \textbf{iNAS:} This approach divides the depth of the network into stacks, each formed by a certain number of block stacks. When searching for the best network depth, the method iNAS uses $A_0 = \{A(B, i, j), A(B, [ij/(j+1)]+ 1, j + 1), A(B, i + 1, j)\} $, each time from the three candidate network structures, selects the model with the best performance on the validation set as the current best model $A(B, i, j)$, and then continues to use the above formula to generate candidate networks, and then select the optimal network until the training iterations are reached. In the first block of each stack (except the first stack), the generated network halves the feature map size and doubles the number of channels. In the final part of the generated network is one classification block, which contains an average pooling layer that reduces the spatial dimension to 1*1, and a fully connected classification layer followed by softmax.
\item \textbf{SVI and mean-filed VI:} Both methods use the basic block as a layer of the network, searching for the best depth of the network based on the infinite depth of the search space. The general search strategy is to train the parameter $\lambda_d$ that determines the depth of the network, making the corresponding network achieve better results. The relationship between the parameter $\lambda_d$ and the network depth is as follows, with $\lambda_d$ as the parameter of truncated Poisson distribution, the minimum value of the probability distribution reaching the specified value (the hyperparameter is set to 0.95 in the experiment) is obtained as the current network depth, and the network output is the weighted sum of the predicted output of each layer and their respective probabilities corresponding to the truncated Poisson distribution. Since $\lambda_d$ corresponds to the largest probability value, it can be approximated that the number of layers represented by the integer near it is the best number of layers found. Compared to backbone networks that use a fixed number of layers, infinite depth avoids the performance impact caused by artificially limiting the search space. Unlike iNAS, the networks corresponding to these two methods halve the feature map size and double the number of channels at layers 4 and 9, while for the output layer of each layer of the network, the kernel size of the mean pooling operation contained in it is (4,4) ((3,3) on MNIST). 
\end{itemize}

\textbf{Experimental Settings.} In all experiments we have some basic same hyperparameter settings. For example, we trained all models using stochastic gradient descent (SGD) with a batch size of 128 and momentum of 0.9. For this optimizer, we use an initial learning rate of 0.01, with a technique called Cosine rampdown~\cite{loshchilov2016sgdr} adjusting it per epoch. A weight decay of 1e-4 was used, and two data augmentation was applied containing four pixels translates and horizontal flips. Further detailed experimental parameter settings are given below for depth search and active learning, respectively.
\begin{itemize}
    \item \textbf{Depth Search:} In the depth search related experiments, we used all the training samples of the corresponding dataset for each of the three datasets. Since the complexity of the datasets varies, we adjusted the training epoch accordingly. On MNIST we trained 50 epochs, while on CIFAR-10 and CIFAR-100 we trained 150 epochs and 200 epochs, respectively. Since the search method of iNAS is to test a fixed three network architectures per active learning cycle, we performed 10 active learning cycles with every candidate network being trained 10 epochs per cycle to achieve the purpose of depth search using all training samples. 
    \item \textbf{Active Learning:} In the active learning related experiments, we used Random Sampling, Coreset and Uncertainty Entropy three querying strategies. And for each active learning experiment, we set the active learning cycle to 12. Then for different datasets and search methods, our parameter settings had some adjustments.  \\
    \textbf{Datasets.} For MNIST dataset, since our network architecture is relatively complex for this task, the initial number of samples for model training was 100, and the amount of data queried per active learning cycle was 200. For the CIFAR-10 and CIFAR-100 datasets, we used an initial sample size of 2000 and a query number of 2000 each cycle, and when the total number of training samples reaches 10000, we updated the query number to 5000. \\
    \textbf{Search methods.} For both types of experiments on fixed network structures and Active-iNAS, we trained for 50 epochs per active learning cycle. We used Resnet-18 and Resnet-34 as our fixed architectures and search space in the range A(Br, 1, 1) to A(Br, 12, 5) was used in the method of Active-iNAS. For experiments on Active-SNDS, we updated the network depth related parameters with a separate optimizer, this depth optimizer also used stochastic gradient descent (SGD) with an initial learning rate 0.05 and when training epochs became 2/3 times the total number of current epochs, the learning rate was adjusted to 0.03. For experiments on mean-field VI, we used the same depth optimizer, but the corresponding learning rate was fixed.
\end{itemize}

\begin{figure}[ht]
	\vskip 0.2in
	\begin{center}
            \centerline{\includegraphics[width=\columnwidth]{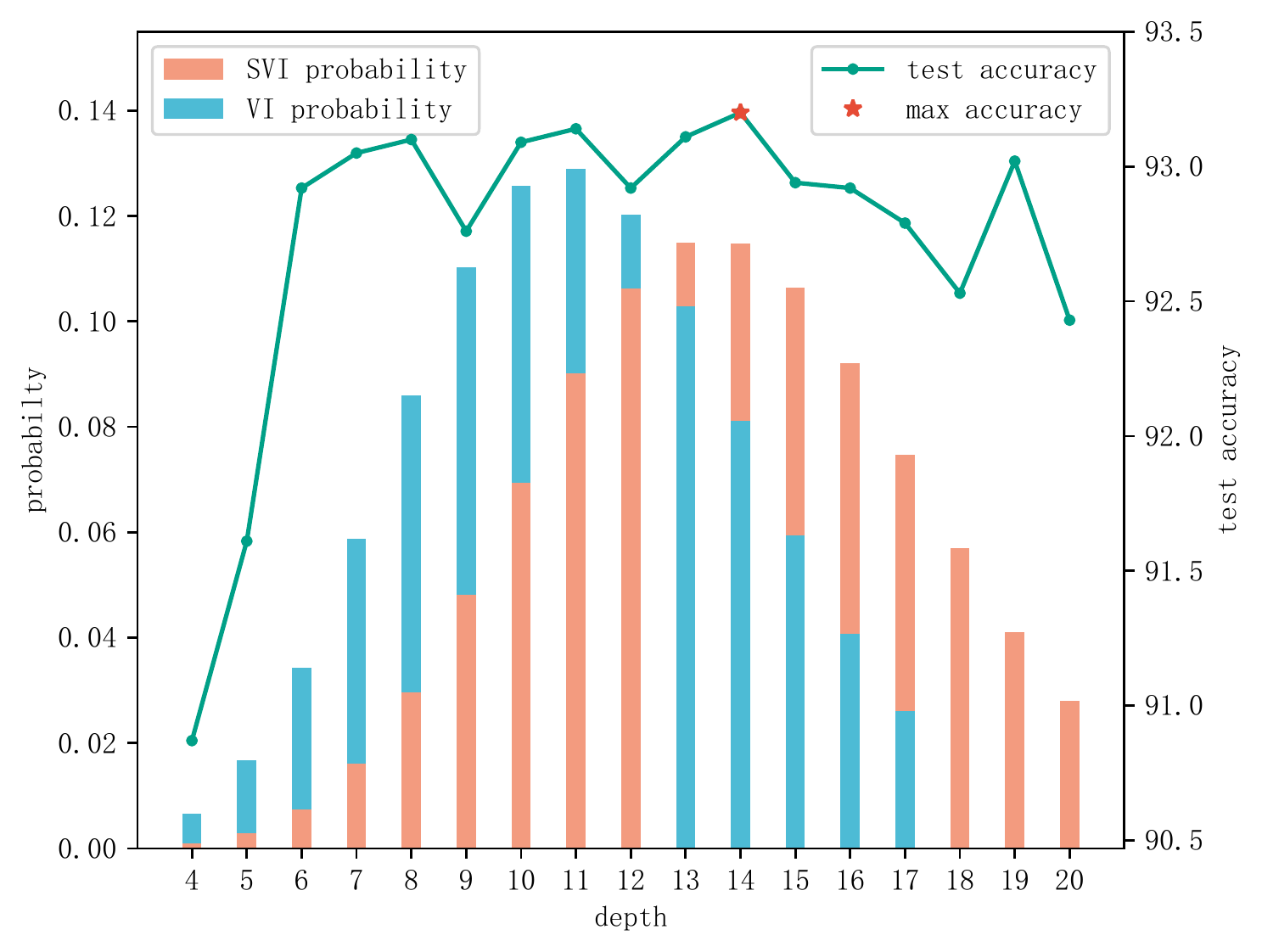}}
            \caption{ Depth search results on CIFAR-10 using methods SVI and mean-field VI. Curve in green – accuracy used fixed networks with depth in the range of 4 to 20, bar in orange – probability for every layer of the depth selected by SVI and bar in blue – probability for every layer of the depth selected mean-field VI. }
            \label{DS_pdf}
	\end{center}
	\vskip -0.2in
\end{figure}

\begin{table*}[ht!]
\caption{Depth results under all training samples on CIFAR-10, MNIST and CIFAR-100. }\label{table:1}
\begin{center}
\begin{small}
\begin{sc}
\renewcommand\arraystretch{1.1}
\setlength{\tabcolsep}{1.5mm}{
\begin{tabular}{lcccccccr}
\hline
DATASET    & ARCHITECTURE & Resnet-18 & Resnet-34 & iNAS & mean-field VI & SVI              \\ \hline
CIFAR-10    & accuracy & 93.10\% & 92.87\% & 84.37\% & 92.83\% & 92.93\%                                     \\ 
                    & depth ($\lambda_d$/structure) & 8 & 16 & 14 ({(}B,7,2{)}) & 17 (10.26) & 20 (12.98)            \\ \hline
MNIST        & accuracy & 98.92\% & 99.15\% & 98.61\% & 99.25\% & 99.16\%                                       \\ 
                    & depth ($\lambda_d$/structure) & 8 & 16 & 14 ({(}B,7,2{)}) & 12 (6.68) & 13 (7.25)                  \\ \hline
CIFAR-100 & accuracy & 55.32\% & 56.64\% & 55.31\% & 70.75\% & 73.19\%                                         \\ 
                    & depth ($\lambda_d$/structure) & 8 & 16 & 12 ({(}B,3,3{)}) & 21 (14.04) & 25 (16.63)               \\ \hline
\end{tabular}}
\end{sc}
\end{small}
\end{center}
\end{table*}

\begin{figure*}[h]
\begin{center}
\centerline{\includegraphics[width=2\columnwidth]{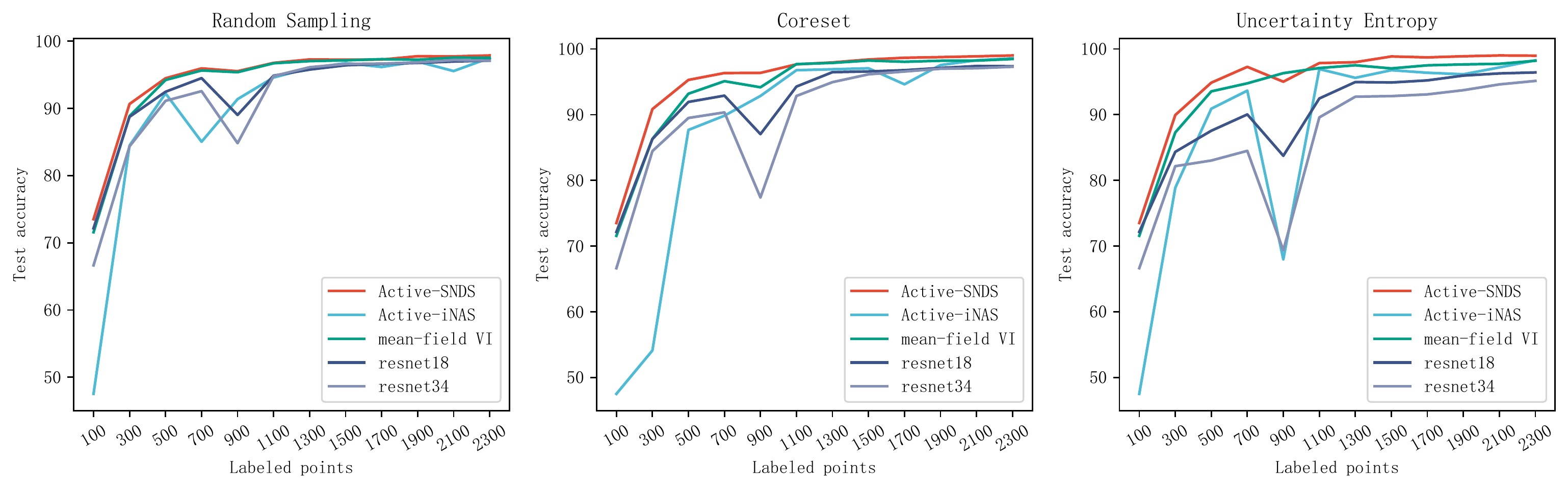}}
\caption{ Active learning curves for MNIST dataset using various query functions. }
\label{result:2}
\end{center}
\end{figure*}

\begin{figure*}[h]
\begin{center}
\centerline{\includegraphics[width=2\columnwidth]{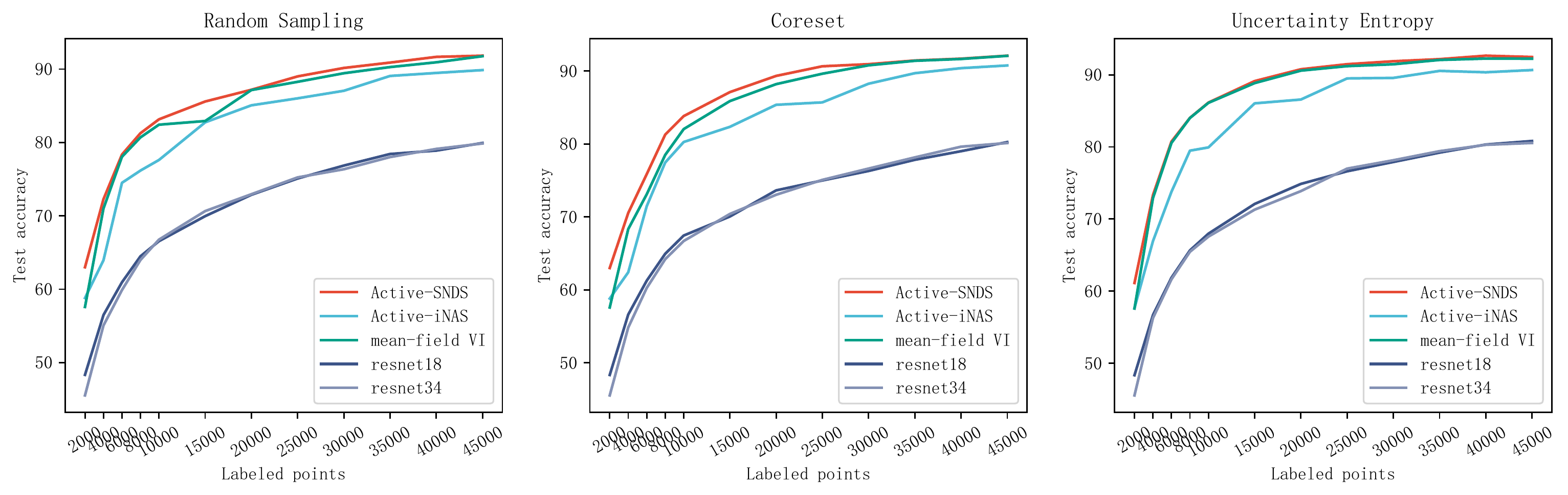}}
\caption{ Active learning curves for CIFAR-10 dataset using various query functions. }
\label{result:3}
\end{center}
\end{figure*}

\begin{figure*}[h]
\begin{center}
\centerline{\includegraphics[width=2\columnwidth]{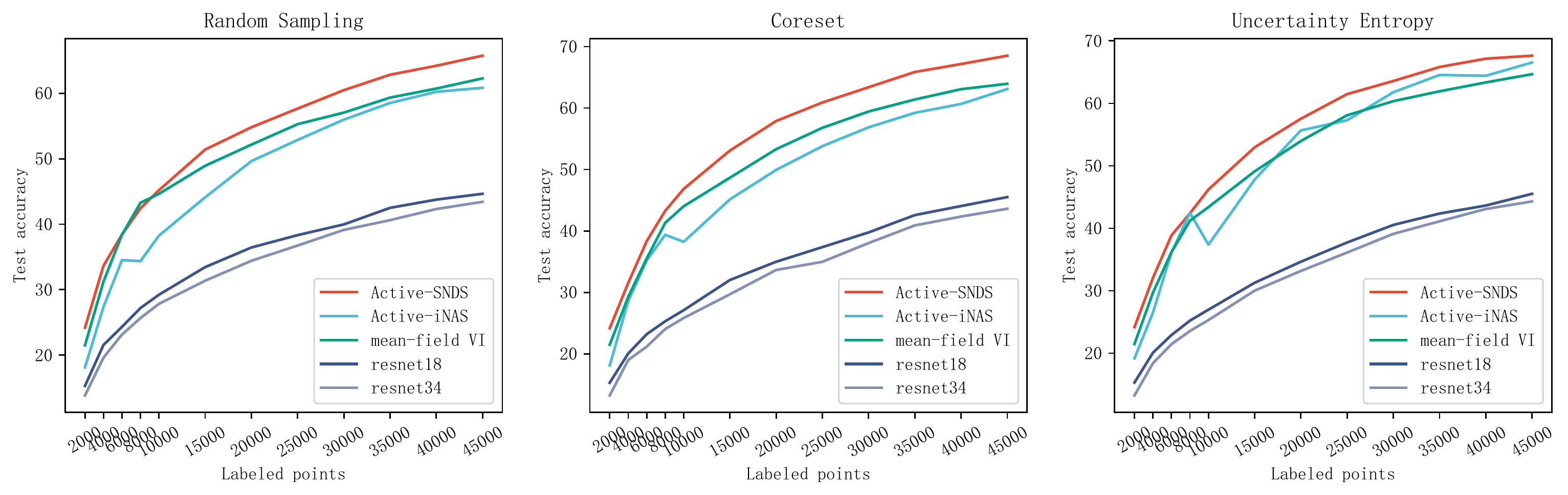}}
\caption{ Active learning curves for CIAFR-100 dataset using various query functions. }
\label{result:4}
\end{center}
\end{figure*}

\subsection{Comparison of depth search experiment results}
In this subsection, we compare the depth search results of SVI, mean-field VI and iNAS, as well as the performance of these three methods and Resnet-18 and Resnet-34 using all datasets on three datasets: MNIST, CIFAR-10 and CIFAR-100. 
Taking CIFAR-10 as an example, we can see the results of the CIFAR-10 shown in Fig. \ref{DS_pdf} and Table \ref{table:1}, the green curve represents the fixed number of layers network around 13, 14 layers presents the highest accuracy, and our proposed method SVI searches the value of parameter $\lambda_d$ is 12.98, close to the optimal number of traversal layers, and compared with the results of mean-field VI search, it can be seen that its value of $\lambda_d$ of 10.26 is significantly less than the optimal depth, so we can know that SVI effectively alleviates the rich-get-richer problem. 

Combining the results of the MNIST and CIFAR-100 in Table \ref{table:1}, we can see that our method has a shallower number of search layers on MNIST and a deeper number of search layers on CIFAR-100, which also corresponds to the actual task complexity. On MNIST, the accuracy of SVI is slightly higher than Resnet-18, Resnet-34 and iNAS, but slightly lower than mean-field VI, because the classification task on the MNIST dataset is simple, and all network structures in the experiment are sufficient to obtain better training results. And the reason for the poor performance of the iNAS method on CIFAR-10 and CIFAR-100 is that the method starts from the shallow layer of search, the search range of each time is limited, and the network parameters of the early training cannot participate in the later training process, so the performance is very poor when the training cycle is insufficient.

\begin{table*}[ht!]
\caption{Depth search results per AL cycle on CIFAR-10 using the query strategy of random sampling. $\lambda_d$ is the parameter that controls the depth in network training and determines the value of depth. }\label{table:2}
\begin{center}
\begin{Large}
\begin{sc}
\renewcommand\arraystretch{1.5}
\resizebox{2.0\columnwidth}{!}{
\begin{tabular}{lccccccccccccccccccccccr}
\hline
Method        & Cycle       & 1       & 2       & 3       & 4       & 5       & 6       & 7       & 8       & 9       & 10      & 11      & 12      \\ \hline
Active-iNAS   & structure   & (B,1,2) & (B,2,2) & (B,2,3) & (B,2,4) & (B,2,4) & (B,3,4) & (B,3,4) & (B,3,4) & (B,3,4) & (B,3,4) & (B,4,4) & (B,5,4) \\ 
              & depth       & 2       & 4       & 6       & 8       & 8       & 12      & 12      & 12      & 12      & 12      & 16      & 20      \\ \hline
mean-field VI & $\lambda_d$ & 1.81    & 4.70    & 6.62    & 7.59    & 8.29    & 8.84    & 9.27    & 9.59    & 9.99    & 10.27   & 10.72   & 11.12   \\ 
              & depth       & 5       & 10      & 12      & 13      & 14      & 15      & 16      & 16      & 16      & 17      & 17      & 18      \\ \hline
Active-SNDS   & $\lambda_d$ & 2.21    & 4.73    & 6.65    & 7.72    & 8.49    & 9.23    & 10.71   & 11.34   & 11.85   & 12.33   & 12.66   & 13.01   \\ 
              & depth       & 6       & 10      & 12      & 14      & 15      & 15      & 17      & 18      & 19      & 19      & 20      & 20      \\ \hline
\end{tabular}}
\end{sc}
\end{Large}
\end{center}
\end{table*}

\begin{table*}[ht!]
\caption{Depth search results per AL cycle on CIFAR-10 using the query strategy of coreset. }\label{table:3}
\begin{center}
\begin{Large}
\begin{sc}
\renewcommand\arraystretch{1.5}
\resizebox{2.0\columnwidth}{!}{
\begin{tabular}{lccccccccccccccccccccccr}
\hline
Method        & Cycle       & 1       & 2       & 3       & 4       & 5       & 6       & 7       & 8       & 9       & 10      & 11      & 12      \\ \hline
Active-iNAS   & structure   & (B,1,2) & (B,2,2) & (B,3,2) & (B,3,3) & (B,3,3) & (B,4,3) & (B,5,3) & (B,5,3) & (B,5,3) & (B,5,3) & (B,5,3) & (B,5,3) \\ 
              & depth       & 2       & 4       & 6       & 9       & 9       & 12      & 15      & 15      & 15      & 15      & 15      & 15      \\ \hline
mean-field VI & $\lambda_d$ & 1.81    & 4.30    & 6.27    & 7.46    & 8.40    & 9.54    & 10.46   & 10.83   & 11.00   & 11.18   & 11.41   & 11.63   \\ 
              & depth       & 5       & 9       & 12      & 13      & 14      & 16      & 17      & 18      & 18      & 18      & 18      & 18      \\ \hline
Active-SNDS   & $\lambda_d$ & 2.21    & 4.70    & 6.58    & 7.95    & 8.90    & 9.91    & 10.49   & 11.50   & 12.15   & 12.58   & 12.82   & 13.11   \\ 
              & depth       & 6       & 10      & 12      & 14      & 15      & 16      & 17      & 18      & 19      & 20      & 20      & 20      \\ \hline
\end{tabular}}
\end{sc}
\end{Large}
\end{center}
\end{table*}

\begin{table*}[ht!]
\caption{Depth search results per AL cycle on CIFAR-10 using the query strategy of uncertainty entropy. }\label{table:4}
\begin{center}
\begin{Large}
\begin{sc}
\renewcommand\arraystretch{1.5}
\resizebox{2.0\columnwidth}{!}{
\begin{tabular}{lccccccccccccccccccccccr}
\hline
Method        & Cycle       & 1       & 2       & 3       & 4       & 5       & 6       & 7       & 8       & 9       & 10      & 11      & 12      \\ \hline
Active-iNAS   & structure   & (B,1,2) & (B,2,2) & (B,2,3) & (B,2,3) & (B,3,3) & (B,4,3) & (B,4,3) & (B,4,3) & (B,5,3) & (B,5,3) & (B,6,3) & (B,7,3) \\
              & depth       & 2       & 4       & 6       & 6       & 9       & 12      & 12      & 12      & 15      & 15      & 18      & 21      \\ \hline
mean-field VI & $\lambda_d$ & 1.81    & 5.15    & 7.41    & 8.64    & 9.74    & 10.40   & 10.85   & 11.04   & 11.16   & 11.29   & 11.49   & 11.67   \\ 
              & depth       & 5       & 10      & 13      & 15      & 16      & 17      & 18      & 18      & 18      & 18      & 18      & 19      \\ \hline
Active-SNDS   & $\lambda_d$ & 2.18    & 5.30    & 7.65    & 8.97    & 9.76    & 10.96   & 11.81   & 12.44   & 12.91   & 13.22   & 13.38   & 13.56   \\ 
              & depth       & 6       & 10      & 13      & 15      & 16      & 18      & 19      & 19      & 20      & 20      & 21      & 21      \\ \hline
\end{tabular}}
\end{sc}
\end{Large}
\end{center}
\end{table*}

\subsection{Comparison of active learning experiment results}
The results of active learning algorithms are usually represented by a curve between the amount of data and the performance of the training samples. We use accuracy in our experiments to represent the performance of the model. For example, in Fig. \ref{result:2} we see the results obtained by three search methods and two fixed architectures for classifying MNIST images using the three querying strategies (Random Sampling, Coreset and Uncertainty Entropy). In red, we see the curve for Active-SNDS. The results of Active-iNAS, mean-field VI, Resnet-18 and Resnet-34 appear in light blue, green, dark blue
and light purple, respectively. The X-axis corresponds to the labeled points consumed, starting from k = 100 (the initial seed size), and ending with 2300 for MNIST (starting from k = 2000 and ending with 45000 for CIFAR tasks).

We show our experimental results for MNIST, CIFAR-10 and CIFAR-100 in Fig. \ref{result:2}, Fig. \ref{result:3} and Fig. \ref{result:4}. We first analyze the results for CIFAR-10 (Fig. \ref{result:3}). Obviously, for all three query strategies, the performance of Active-SNDS is optimal throughout the entire range. 

At the same time, as shown in Table \ref{table:2}, \ref{table:3} and \ref{table:4}, from the comparison of the number of network layers searched by Active-SNDS and their parameters and the number of network layers searched by Active-iNAS, it can be seen that the actual number of layers used by Active-SNDS will be less than that of Active-iNAS, that is, the value of parameter $\lambda_d$ is smaller because Active-SNDS continues to train deeper networks on the basis of the previous shallow network, so in the case of few training epochs, Active-SNDS can better train the model to get better performance; but it can also be seen that in Active-SNDS, the actual network depth is deeper because we do not limit the search space, so the depth of the backbone network can be increased infinitely according to the actual situation. And comparing the search results of Active-SNDS and mean-field VI, we can see that Active-SNDS can search for a deeper depth and obtain better results, further verifying the alleviation of Active-SNDS to the rich-get-richer problem.

Since we used the basic block as a layer, we can think of Resnet-18 and Resnet-34 as layer 8 and layer 16 networks, respectively. So, combining the number of network layers and the performance of the five types of experiments per cycle, we can know that, Active-SNDS tends to generate simple networks with fewer layers in the early stage when there are few training data and increase network complexity in the later stage with the number of training data increasing.

We now see the results of MNIST~(Fig. \ref{result:2}) and CIFAR-100~(Fig. \ref{result:4}) experiments. Since the task on MNIST is well-known simple, all the curves in Fig. \ref{result:2} show good performance, but the advantages of Active-SNDS can still be reflected in the figure, and another interesting point is that we only need a smaller amount of data to achieve an almost best performance effect. The classification task on CIFAR-100 is the opposite of MNIST, and the more complex classification task makes network training harder, so we can see that the overall accuracy has decreased, and we can also see that the performance of our method is more obvious than that of other network architectures. Since Resnet-18 and Resnet-34 train the same 50 epochs in each active learning cycle, and such fewer training cycles result in the more complex Resnet-34 not achieving better results than the simpler Resnet-18, the results presented in Fig. \ref{result:4} occurs.

\section{Conclusion}
In this paper, we propose a novel active learning strategy Active-SNDS and an efficient neural depth search method SVI. We theoretically demonstrate that the mean-field assumption of previous VI-based methods can cause the rich-get-richer problem and restore it with SVI. Experimental results over enormous datasets and acquire methods show that SVI has better incremental network depth search performance and Active-SNDS outperforms Active-iNAS and the direct VI-based learning method in active learning processes. In summary, our proposed methods demonstrate significant improvements in deep active learning.

\bibliographystyle{IEEEtran}
\bibliography{main}{}

\newpage

\newpage

\end{document}